\documentclass[authoryear,review,12pt]{elsarticle}

\makeatletter
\def\ps@pprintTitle{}
\makeatother

\usepackage{amsmath}
\usepackage{xcolor}
\usepackage{multirow}
\usepackage{lineno}
\usepackage{graphicx}
\usepackage{adjustbox}
\usepackage{multirow}%
\usepackage{amsmath,amssymb,amsfonts}%
\usepackage{amsthm}%
\usepackage{mathrsfs}%
\usepackage[title]{appendix}%
\usepackage{textcomp}%
\usepackage{manyfoot}%
\usepackage{booktabs}%
\usepackage{algorithm}%
\usepackage{algorithmicx}%
\usepackage{algpseudocode}%
\usepackage{listings}%
\usepackage{pdflscape}
\usepackage{natbib}
\setcitestyle{authoryear,open={(},close={)}}

\usepackage{tikz}
\usetikzlibrary{tikzmark,arrows.meta}
\usepackage{geometry}
\graphicspath{{fig/}}

\newcommand{\revb}[1]{{\color{black}{#1}}}
\usepackage{CJKutf8}
\usepackage[colorlinks=true, citecolor=blue]{hyperref}

\begin{document}

\begin{frontmatter}


\title{Mapping the Vanishing and Transformation of Urban Villages in China}

\author[aff1,aff2]{Wenyu Zhang} 
\author[aff3]{Yao Tong}
\author[aff1]{Yiqiu Liu}
\author[aff1]{Rui Cao\corref{cor1}}
\ead{ruicao@hkust-gz.edu.cn}
\cortext[cor1]{Corresponding author}

\affiliation[aff1]{organization={Thrust of Urban Governance and Design, Society Hub},addressline={The Hong Kong University of Science and Technology (Guangzhou)}, city={Guangzhou},country={China}}
\affiliation[aff2]{organization={Department of Geography}, addressline={National University of Singapore}, city={Singapore}, country={Singapore}
} 
\affiliation[aff3]{organization={School of Architecture}, addressline={Tsinghua University}, city={Beijing}, country={China}
} 

\begin{abstract}
Urban villages (UVs), informal settlements embedded within China’s urban fabric, have undergone widespread demolition and redevelopment in recent decades. However, there remains a lack of systematic evaluation of whether the demolished land has been effectively reused, raising concerns about the efficacy and sustainability of current redevelopment practices. To address the gap, this study proposes a deep learning–based framework to monitor the spatiotemporal changes of UVs in China. Specifically, semantic segmentation of multi-temporal remote sensing imagery is first used to map evolving UV boundaries, and then post-demolition land use is classified into six categories based on the ``remained–demolished–redeveloped'' phase: incomplete demolition, vacant land, construction sites, buildings, green spaces, and others. Four representative cities from China’s four economic regions were selected as the study areas, i.e., Guangzhou (East), Zhengzhou (Central), Xi’an (West), and Harbin (Northeast). The results indicate: 1) UV redevelopment processes were frequently prolonged; 2) redevelopment transitions primarily occurred in peripheral areas, whereas urban cores remained relatively stable; and 3) three spatiotemporal transformation pathways, i.e., synchronized redevelopment, delayed redevelopment, and gradual optimization, were revealed. This study highlights the fragmented, complex and nonlinear nature of UV redevelopment, underscoring the need for tiered and context-sensitive planning strategies. 
By linking spatial dynamics with the context of redevelopment policies, the findings offer valuable empirical insights that support more inclusive, efficient, and sustainable urban renewal, while also contributing to a broader global understanding of informal settlement transformations.
\end{abstract}


\end{frontmatter}

\section{Introduction}
\label{sec1}
With the rapid acceleration of global urbanization, informal settlements have become a critical challenge for sustainable urban development. Known internationally as slums \citep{LI2023104863}, favelas \citep{carvalho2021beyond}, or shantytowns \citep{LI2018106}, these areas are typically characterized by informal construction, inadequate infrastructure, and unclear legal status \citep{un-habitat2004}. The United Nations' Sustainable Development Goals (SDG) emphasize the need to improve the living conditions of residents in such settlements and promote their integration into formal urban systems \citep{un2023rescuing}. Achieving this requires accurate mapping of their spatial distribution and temporal evolution \citep{BREUER2024103056}.

In the context of China’s urbanization, informal settlements are primarily represented by urban villages (UVs)—rural enclaves embedded within expanding urban built-up areas \citep{liuUrbanVillagesChina2010,cao2025mapping}. Rooted in the persistence of collective rural land ownership, UVs have long existed outside formal urban planning and governance frameworks, resulting in unregulated growth and inadequate public services \citep{chenHierarchicalApproachFinegrained2022,tu_towards_2024}. Combined with high population mobility and fragmented socio-spatial structures, these conditions have given rise to persistent informality, functional disorder, and heightened social vulnerability \citep{cao2025mapping}. Despite these challenges, UVs play a critical role in providing affordable housing for migrant workers and low-income groups \citep{jin_destigmatizing_2024}, making their governance crucial for promoting urban equity, social inclusion, and sustainable development. 

Since the early 2000s, large-scale UV redevelopment began in China, primarily through demolition and real estate-led reconstruction to enhance land value \citep{wu_ripples_2025}. Although this model has yielded short-term economic gains and visual improvements, systematic evaluations of post-demolition land-use transformation and utilization efficiency remain limited. Moreover, concerns over long-term sustainability, equity, and social inclusion have sparked growing criticism of this economically driven model \citep{li_dawn_2021, chen_arrival_2023}. In recent years, as China’s urbanization enters a phase of stock-based renewal, UV redevelopment has not only regained policy attention but also expanded into a nationwide agenda \footnote{The State Council of the People's Republic of China. (2024, November 15). Policy Support for Urban Village Redevelopment Expanded. \url{https://www.gov.cn/lianbo/bumen/202411/content_6987166.htm}}. New strategies now prioritize livelihood improvement, moderate demolition, and context-sensitive interventions to stabilize the housing market and support sustainable transformation \citep{SHEN2021105571,Ye2024CityPlanningReview}. However, despite the scope and speed of these transformations, a fundamental knowledge gap remains: we still know relatively little about what happens to UVs after demolition. Are they swiftly rebuilt into new residential or public services? Are they left vacant? These post-demolition trajectories critically affect housing equity, spatial efficiency, and redevelopment success, yet they are seldom systematically monitored or evaluated. This gap poses urgent challenges for both academic research and urban governance, underscoring the need for scalable and evidence-based approaches to monitoring UV demolition and redevelopment.

\subsection{Related work in mapping and monitoring urban villages}
\label{sec 1.1}
Advances in geospatial technologies and artificial intelligence (AI) have enabled the study of UVs using multi-source spatiotemporal data \citep{caoResponsibleUrbanIntelligence2023,yue2025shaping}. Compared with traditional land surveys, deep learning methods can extract complex spatial features from large-scale remote sensing imagery and support dynamic change detection and analysis across broad spatial and temporal ranges. Despite these promising developments, current research still faces major challenges: 1) difficulties in developing standardized mapping methods, 2) limited efforts in long-term and cross-regional monitoring, and 3) insufficient understanding of post-demolition transformation and redevelopment outcomes \citep{cao2025mapping}.  

The first challenge lies in the spatial heterogeneity of UVs and the resulting difficulty in developing standardized mapping methods. Accurately mapping the spatial extent and temporal transformation of UVs is foundational to understanding their evolutionary dynamics and supporting effective redevelopment \citep{cao2025mapping}. Current UV mapping relies heavily on very-high-resolution remote sensing (VHR-RS) imagery (e.g., Google Earth, Gaofen-2), often with spatial resolutions below 3 meters, and in some cases reaching sub-meter levels \citep{gao_novel_2024,chenUVLensUrbanVillage2021}. To reflect heterogeneity and functional characteristics, researchers increasingly incorporate socially sensed data, such as street view imagery, points of interest (POIs), mobile phone signals, and shared bike trajectories \citep{chenHierarchicalApproachFinegrained2022,liMappingUrbanVillages2023,tu_towards_2024}. While these data enrich functional understanding, cross-city applications remain hindered by high acquisition costs, inconsistent quality, and the absence of standardized data fusion protocols. In addition, the vast geographic span of China gives rise to significant regional variations in UV morphology (as illustrated in the study area example in Figure \ref{fig:characteristics}), further complicating efforts to develop a one-size-fits-all identification strategy. Against this backdrop, VHR-RS imagery combined with deep learning has become a more feasible and unified pathway for scalable and temporally consistent UV monitoring. 

The second challenge stems from the limited spatiotemporal scope of most existing UV studies. Early efforts typically focused on static, single-city mapping of UVs using manual or rule-based methods. With the increasing availability of VHR-RS imagery and the evolution of deep learning models, recent work has moved toward more automated and scalable approaches. Convolutional Neural Networks (CNNs) and Transformer architectures now enable end-to-end recognition with enhanced precision \citep{weiMonitoringUrbanVillages2015,panDeepLearningSegmentation2020,fanFineScaleUrbanInformal2022,zhangUVSAMAdaptingSegment2024}.
However, the spatial application of these models has largely remained concentrated in very limited cities with well-developed data infrastructures, such as Shenzhen, Guangzhou, and Beijing \citep{cao2025mapping}. As methodological systems matured and data coverage expanded, only recently have studies begun expanding the study area \citep{chai_fine-grained_2024,fan_refined_2025,tu_towards_2024,10888896,lin_long-term_2024}. Temporally, UV mapping is also evolving from single-year snapshots toward dynamic multi-year monitoring \citep{huang_spatiotemporal_2015,lin_long-term_2024}. 
Despite recent advances, existing studies remain limited in short-term and city-specific analyses, lacking systematic examination across broader spatial and temporal scales.

The third challenge is the lack of fine-grained assessments of land use transition and redevelopment efficiency following UV demolition. While earlier study has described general patterns of UV transformation at the district level \citep{weiVanishingRenewalLandscape2023}, few have interpreted land reuse pathways or evaluated actual outcomes at a broader scale. Such detailed assessments are technically demanding, as demolition and redevelopment often progress asynchronously, producing transitional landscapes with blurred boundaries and mixed-use patterns that are difficult to classify \citep{cao2025mapping}. In addition, inconsistencies in the quality of VHR-RS imagery—caused by seasonal variability, uneven lighting, and occlusions from shadows or clouds—further complicate analysis \citep{rolf_mission_2024,wen2021change}.

To address these challenges, a dynamic monitoring framework is needed to capture the full transformation cycle, from demolition to redevelopment, while maintaining both semantic clarity and temporal adaptability. In response, this study develops a deep learning framework using VHR-RS imagery to map UVs over time and monitor their post-redevelopment land-use transitions across regions. Specifically, we train deep learning models on baseline imagery and apply it to subsequent years to ensure spatial consistency. The resulting change areas are further attributed to functional types using POIs and OpenStreetMap (OSM) data. Without requiring multi-temporal training or domain adaptation, this framework achieves a balance between spatial accuracy and meaningful functional interpretation. It provides a scalable and cost-efficient solution for long-term and cross-region monitoring of UV transformation, enabling more informed and adaptive urban planning. To reflect regional diversity and variations in redevelopment practices, four major cities representing China's four economic divisions are selected: Guangzhou (East), Zhengzhou (Central), Xi’an (West), and Harbin (Northeast). 

The main contributions of this study can be summarized as follows:
\begin{itemize}
    \item We propose a deep learning-based framework for dynamic mapping and spatiotemporal monitoring of UVs, enabling scalable, cross-regional, and long-term monitoring of UV transformation;
    \item To the best of our knowledge, this is the first study to systematically quantify the vanishing and transformation of UVs across multiple Chinese mega-cities, mapping their spatial restructuring patterns and revealing redevelopment efficiency;
    \item Through a comparative analysis, we identify three distinct post-demolition land-use transformation patterns: synchronized redevelopment, delayed redevelopment, and gradual optimization. This typology reveals the fragmented and non-linear nature of UV transformation, and offers empirical evidence to inform differentiated planning strategies and policy-making for informal settlement governance. 
\end{itemize}

The remainder of this paper is organized as follows. Section \ref{sec2} introduces the study area and data. Section \ref{sec3} presents the proposed framework for UV spatiotemporal change analysis. Section \ref{sec4} reports and analyzes the results, offering both qualitative and quantitative insights into the spatiotemporal dynamics of UVs. Section \ref{sec5} discusses research limitations and policy implications. Section \ref{sec6} concludes the paper with the main contributions and findings. 

\section{Study area and data}
\label{sec2}
\subsection{Study area }
\label{sec2.1}
\begin{figure}
    \centering
    \includegraphics[width=1 \linewidth]{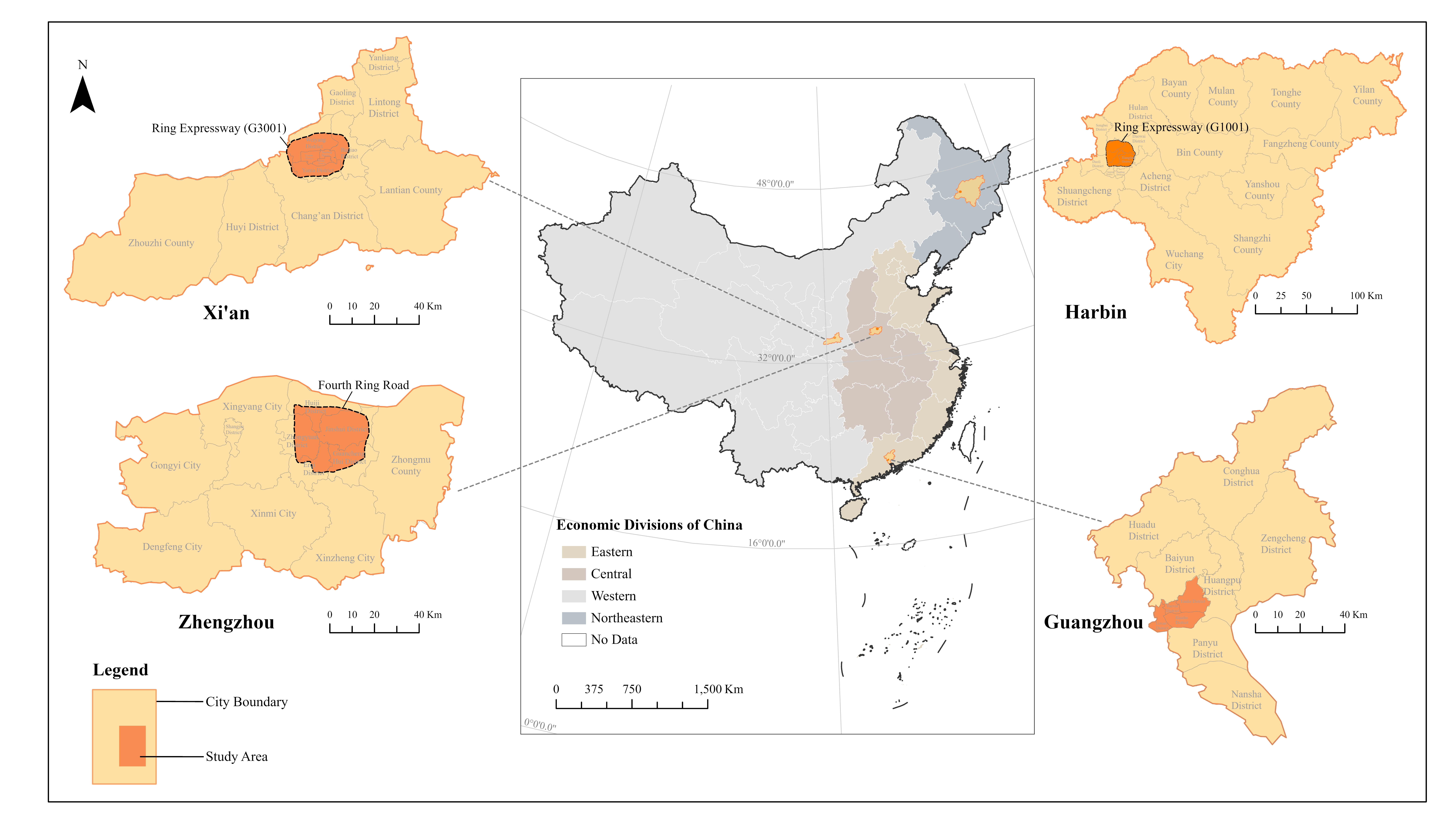}
    \caption{Study area. Four cities from China's four economic divisions are selected, including Guangzhou (East), Zhengzhou (Central), Xi’an (West), and Harbin (Northeast). \revb{The major ring roads are indicated with dashed lines, including the Fourth Ring Road in Zhengzhou, the Xi’an Ring Expressway (G3001, national expressway code), and the Harbin Ring Expressway (G1001, national expressway code).}} \label{fig:StudyArea}
\end{figure}

Rapid urban expansion in China has driven the large-scale emergence and evolution of UVs, marked by spatial, functional, and social transformations shaped by diverse socioeconomic and developmental contexts \citep{liMappingUrbanVillages2023}. Despite common origins, their trajectories diverge notably across regions due to varying structures and strategies. This study selects Guangzhou, Zhengzhou, Xi’an, and Harbin as the study areas (see Figure \ref{fig:StudyArea}), based on the following criteria: 1) Representativeness: All four cities are rapidly developing provincial capitals with pronounced UV issues, spanning across different city tiers, including first-tier, new first-tier, and second-tier; 2) Regional coverage: The selected cities represent China’s four major economic regions \footnote{National Bureau of Statistics of China (2011). Major macro-regional economic divisions. \url{https://www.stats.gov.cn/zt_18555/zthd/sjtjr/dejtjkfr/tjkp/202302/t20230216_1909741.htm}}, i.e., East (Guangzhou), Central (Zhengzhou), West (Xi’an), and Northeast (Harbin); 3) Diversity: The cities include old industrial cities, transport hub cities, and emerging urban centers shaped by post-reform development, allowing for comparison across different urbanization trajectories; 4) Redevelopment stages: The four cities vary in their timing, pace, and intensity of UV redevelopment. 

\begin{figure}
    \centering
    \includegraphics[width=1\linewidth]{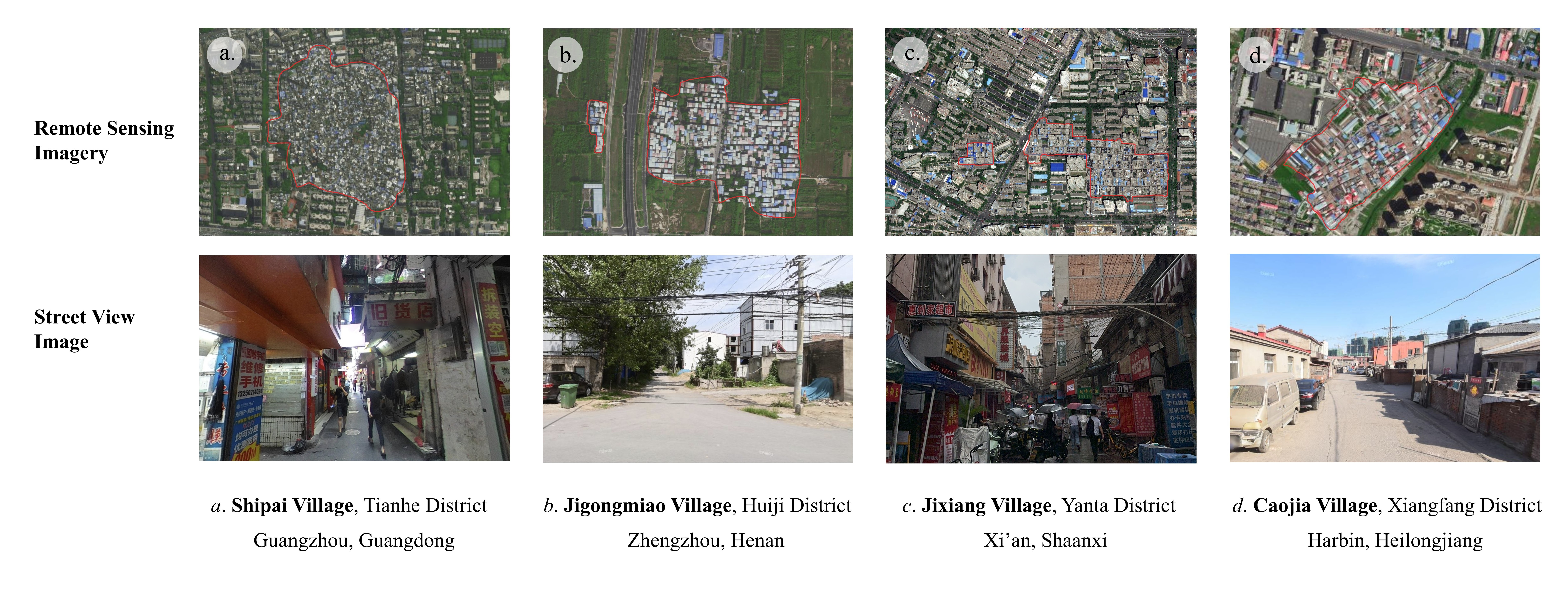}
    \caption{Representative urban village types in the study area. (a) High-density area with blurred boundaries and active commerce; (b) Suburban village with clear edges and traditional structure; (c) Compact, independent core-area village with dense commercial activity; (d) Loosely organized settlement with semi-urban characteristics. The red boundaries of UVs in the remote sensing imagery are delineated manually. (Data source: Baidu Map (as of July 2024))
}
    \label{fig:characteristics}
\end{figure}

In delineating the study areas, this research focuses on the core built-up zones of each city and further refines the boundaries based on administrative divisions and urban development contexts. Specifically, Guangzhou's study area includes the four central districts: Liwan, Yuexiu, Haizhu, and Tianhe. For Zhengzhou, the area within the Fourth Ring Road is selected, encompassing Zhongyuan, Erqi, Jinshui, Guancheng Huizu, and Huiji districts. Xi’an and Harbin are defined by their respective ring expressways. Xi’an's study area includes the core urban districts of Beilin, Xincheng, Lianhu, Weiyang, Yanta, and parts of Baqiao, while Harbin’s area includes Daoli, Nangang, Daowai, and Xiangfang districts. To further contextualize spatial heterogeneity, typical UVs are visually examined using remote sensing and street view imagery (see Figure \ref{fig:characteristics}), revealing notable differences in spatial boundaries, surrounding environments, and building forms.

\subsection{Data sources}
\label{sec2.2}
This study integrates multi-source data to analyze the spatial distribution of UVs, land use change, and assess the spatial impacts of redevelopment. As summarized in Table \ref{tab:dataset}, the data sources include VHR-RS imagery, village name, road network, and POI data.
\begin{table}
    \centering
    \caption{Overview of datasets.}
    \label{tab:dataset}
    \resizebox{\textwidth}{!}{ 
    \begin{tabular}{p{4cm}p{6cm}p{3cm} } 
        \toprule
        
        Data & Year (City)& Source \\
        
        \midrule
        \multirow[t]{4}{*}{\parbox[t]{4cm}{\raggedright Remote sensing imagery}} & 2015, 2019, 2023 (Guangzhou) & \multirow[t]{4}{*}{\parbox[t]{3cm}{\raggedright Google Earth (approximately 0.5 m resolution)}}\\
         & 2009, 2016, 2022 (Zhengzhou)& \\
        & 2015, 2019, 2023 (Xi’an)& \\
        & 2010, 2016, 2022 (Harbin)& \\
        \multirow[t]{2}{*}{Village name}& 2023 (Guangzhou, Xi’an)& \multirow[t]{2}{*}{Government}\\
 & 2022 (Zhengzhou, Harbin)&\\
        
        \multirow[t]{2}{*}{Road network}& 2019, 2023 (Guangzhou, Xi’an)& \multirow[t]{2}{*}{OpenStreetMap}\\
 & 2016, 2022 (Zhengzhou, Harbin)&\\
        \multirow[t]{2}{*}{POIs}& 2019, 2023 (Guangzhou, Xi’an)&\multirow[t]{2}{*}{Amap}\\
        
        &  2016, 2022 (Zhengzhou, Harbin)& \\
        
        \bottomrule
        
    \end{tabular}
    } 
\end{table}
\revb{VHR-RS imagery from Google Earth (approximately 0.5 m resolution) was used for mapping UV boundaries and detecting temporal changes. Imagery years were chosen based on two criteria: major redevelopment policy milestones (see Section \ref{sec5.3}) and image availability and quality. Although remote sensing imagery often contains clouds and building shadows, we ensured that the selected images were clear overall and that the affected areas did not overlap with UVs, thereby minimizing potential quality issues in subsequent analysis.} Two time-series spans were constructed: 2015, 2019, and 2023 for Guangzhou and Xi’an (8 years); 2009/2010, 2016, and 2022 for Zhengzhou and Harbin (13 years). We also used the catalog of village names, consolidated from local government public information and field research, \revb{to assist in the labeling of the training data for deep learning models}. Road network data from OSM, including highways and urban roads, were employed to delineate study boundaries and analyze spatial patterns. Socially sensed data such as POIs has proven valuable for attributing functional meaning to land-use transitions \citep{liu_review_2022}. In this study, POI data from Amap (a.k.a. Gaode Maps), covering commercial, residential, educational, medical, and transport categories, support functional classification \citep{tu_towards_2024,liMappingUrbanVillages2023}. Based on the POIs attributes, we constructed a redevelopment classification to capture land-use transformation outcomes.

\section{Methodology}
\label{sec3}
\begin{figure}
    \centering
    \includegraphics[width=1\linewidth]{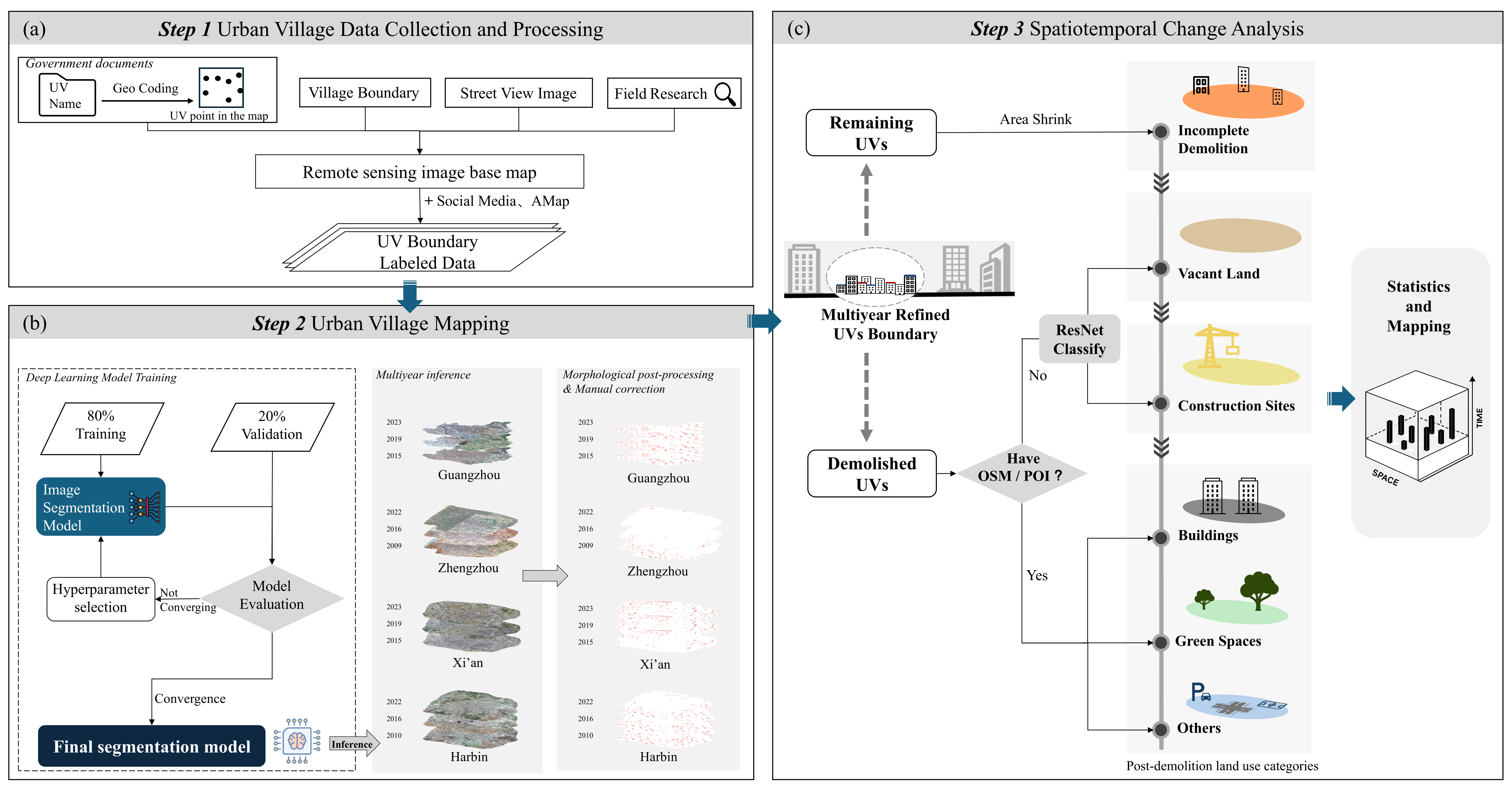}
    \caption{The proposed framework for urban village spatiotemporal change analysis, including three major steps: (a) urban village data collection and preprocessing, (b) urban village mapping, (c) spatiotemporal change analysis.}
    \label{fig:flowchart}
\end{figure}

This section elaborates on the proposed framework for mapping the dynamic spatial extent and land use changes of UVs. As illustrated in Figure \ref{fig:flowchart}, it includes three steps: 1) Data collection and preprocessing. This step integrates multi-source datasets to build a high-quality annotated dataset for deep learning model training. 2) UV mapping. Deep learning–based semantic segmentation models are applied to multi-temporal remote sensing imagery, with model optimization and morphological post-processing used to enhance boundary precision and temporal consistency. 3) Spatiotemporal change analysis. Based on the identified UVs, this step recognizes the land-use types of the demolished UV areas, categorizes transformation pathways, and employs spatial statistics to reveal UV evolution patterns across the study area. The details of each step are elaborated in the following subsections.

\subsection{\revb{Urban village data collection and preparation}}
\label{sec3.1}
\revb{The foundation of the framework lies in the construction of a high-quality labeled dataset that can support robust model training and evaluation. This process began with the careful selection of imagery years, where policy relevance, image quality, and sample availability were jointly considered. For Xi’an and Guangzhou, imagery from 2019 was selected as it represented the mid-stage of redevelopment, when UVs remained abundant and morphologically distinct. In Zhengzhou, imagery from 2009 captured the early redevelopment stage before widespread clearance, while in Harbin, imagery from 2010 provided the earliest reliable baseline prior to large-scale demolition. 

Training samples were labeled after the team established a shared understanding of UV characteristics across the four cities through studying and discussing their background and typical forms. Once the temporal baseline was determined, candidate UVs were first located from geocoded government-published name lists with satellite imagery, then refined using street view image data, village boundary maps, and user-contributed place names from Amap. Cases with blurred boundaries or complex morphologies were resolved through group consensus and, when necessary, validated by field checks. Sampling was distributed across subregions (east, west, south, north, central) to capture intra-city variation. Final labeling was independently performed by two members with cross-checking, followed by a third-member review, with each labeler spending approximately 8–10 person-hours per city. 

Given the inherent spatial autocorrelation and temporal dependency of remote sensing data \citep{karasiak2022spatial}, conventional random sampling often inflates accuracy and underestimates generalization error \citep{sun2023spatial,rolf_mission_2024}. To reduce label leakage and mitigate overfitting, we adopted geographical blocking and spatial buffering strategies \citep{ploton2020spatial}. Geographical blocking partitions the study area into non-overlapping units and assigns entire units exclusively to training or validation, while spatial buffering establishes exclusion zone around validation samples and removes nearby training samples, thereby ensuring spatial independence between the two sets \citep{roberts2017cross}. In addition, validation sites were carefully selected to include diverse UV morphologies, such as compact clusters embedded in high-rise districts and fringe settlements adjacent to farmland. }

Finally, imagery was preprocessed into 1024$\times$1024 pixel tiles with 512-pixel overlap to enhance sample diversity and segmentation smoothness. As presented in Table \ref{tab:training_set}, the labeled dataset was divided into training and validation sets with a ratio of 80:20, with a roughly balanced ratio of positive to negative pixels (1:1). \revb{To provide a clearer sense of scale, the total UV area covered by the training and validation samples in each city is also reported.}
\begin{table}
    \centering
    \caption{Overview of the training and validation datasets, including the number of image tiles and the total UV area covered.}
    \label{tab:training_set}
    \begin{tabular}{ccccc} 
        \hline
        City& \# Training Image& \# Validation Image& Sampled UV Area (km²)&Year
\\
        \hline

        Guangzhou& 524& 131& 9.77& 2019
\\
        Zhengzhou& 630& 153& 11.68& 2009
\\
        Xi’an& 1026& 257& 19.13& 2019
\\
        Harbin& 821& 197& 15.18& 2010
\\
        \hline
    \end{tabular}
\end{table}

\subsection{Urban village mapping based on semantic segmentation}
\label{sec3.2}
With the dataset prepared, deep learning-based semantic segmentation models can be trained to map UVs across urban areas. To improve generalizability and precision, we compared three semantic segmentation models widely used for UV mapping across-regions: U-Net, DeepLab-v3+, and UV-SAM. These models represent different segmentation strategies tailored to the unique challenges posed by UVs, such as irregular boundaries, varying scales, and heterogeneous forms:
\begin{itemize}
    \item U-Net \citep{ronneberger2015unet} adopts a symmetric encoder–decoder architecture with skip connections to preserve fine-grained spatial features, making it suitable for high-resolution image segmentation tasks; 
    \item DeepLab-v3+ \citep{chen2018encoderdecoderatrousseparableconvolution} leverages atrous convolution and atrous spatial pyramid pooling (ASPP) to capture multi-scale contextual information, thereby enhancing segmentation robustness in complex environments; 
    \item UV-SAM \citep{zhangUVSAMAdaptingSegment2024} integrates the state-of-the-art SegFormer with SAM to first produce coarse masks and then refine boundaries via SAM for enhanced accuracy.
\end{itemize}

Model training was performed on the labeled dataset, with hyperparameter tuning (e.g., learning rate, batch size, number of epochs) used to optimize convergence and stability. Training minimized the segmentation loss, while validation performance was assessed using standard metrics, including mean Intersection over Union (mIoU), UV-specific IoU (UV-IoU), recall (UV-Recall), and precision (UV-Precision). To further evaluate the cross-domain generalization capability of each model, we additionally constructed a mixed-city dataset by sampling data from the four cities in equal proportions. This dataset was used for both training and validation to simulate cross-domain scenarios, and the mean IoU on the validation split (Mixed regions-mIoU) served as an indicator of model adaptability across heterogeneous urban environments. As the primary goal was accurate boundary delineation, the final model selection was based on performance metrics rather than architectural differences. For each city, the best-performing model was applied to multi-temporal imagery to support large-scale UV mapping. 

To ensure accurate spatiotemporal change detection analysis, a post-processing pipeline was implemented to refine the model outputs. Morphological closing was first applied to fill small gaps and enhance boundary continuity, followed by opening operations to eliminate noise and smooth object contours \citep{chenHierarchicalApproachFinegrained2022}. Next, temporal alignment of UV boundaries was conducted by comparing outputs across years with reference maps to maintain geometric consistency and minimize spatial misalignment caused by temporal offsets, sensor variation, or model bias. Finally, manual verification was conducted to correct false positives and negatives, producing a reliable and consistent UV boundary dataset. 
 
\subsection{Spatiotemporal land use change recognition and analysis}
\label{sec3.3}
To investigate the physical transition and functional restructuring pathways of UVs during urban renewal, this study employs a three-stage lifecycle framework comprising \textit{Remained}, \textit{Demolished}, and \textit{Redeveloped} stages (as shown in Figure \ref{fig:flowchart}(c)). Drawing upon the ``maintain, demolish, and re-purpose" transformation logic proposed by \cite{johnson_maintain_2014}, this study applies a post-classification approach to spatiotemporal change detection, in which multi-temporal images are segmented independently and their classification results are compared at the pixel level to identify changes \citep{shi2020change}. This method incorporates UV boundaries confirmed each year, enabling consistent and interpretable tracking of their lifecycle transitions across space and time. 

The classification primarily relies on UV boundaries extracted from high-resolution remote sensing imagery captured at three observation periods ($T_1$, $T_2$, and $T_3$). Parcels consistently identified as UVs throughout all observation years were classified as Remained. In contrast, parcels detected in earlier periods ($T_1$ or $T_2$) but absent in the most recent period ($T_3$) were categorized as Demolished, with their spatial extents ($DB_{T_3}$) computed as:
\begin{equation}
\label{eq:DB}
DB_{T_3} = (B_{T_1} \cup B_{T_2}) - B_{T_3}
\end{equation}
where $B_{T_1}, B_{T_2}, B_{T_3}$ refers to the spatial extent (boundaries) of detected UVs in study year $T_1, T_2, T_3$, respectively. Further, demolished parcels redeveloped into stable urban functions by the year $T_3$ are classified as Redeveloped. 

To characterize the nuanced transformations following UV demolition, we further subdivide the Demolished and Redeveloped stages into six mutually exclusive land-use categories: Incomplete demolition, Vacant land, Construction sites, Buildings, Green Spaces, and Others (see Table \ref{tab:six_land_use_types}). These land-use categories encompass the entire transformation process from transitional to fully functional states, thereby forming an operational and generalizable spatiotemporal classification system.

\begin{table}[ht]
    \centering
    \caption{Classification system of the post-redevelopment land use of UVs.}
    \label{tab:six_land_use_types}
    \resizebox{\textwidth}{!}{
    \begin{tabular}{p{3cm} p{4cm} p{5cm} p{5cm}}
        \toprule
        \textbf{Transformation Phase} & \textbf{Land Use Type} & \textbf{Description} & \textbf{Identification Method} \\
        \midrule

        \tikzmark{top}Remained & Urban Villages & UVs continuously identified across $T_1$, $T_2$, and $T_3$ & Identified by temporal consistency of UV boundaries \\
        \cline{2-4}

        \multirow{3}{=}{\tikzmark{mid}Demolished}
        & Incomplete Demolition & UVs with significantly reduced area by $T_3$, indicating ongoing but incomplete demolition & Defined by geometric reduction across years (see Equation~\ref{eq:DB}) \\
        \cline{2-4}
        & Vacant Land & Cleared land with no redevelopment, showing exposed soil and low vegetation & Classified using ResNet model \\
        \cline{2-4}
        & Construction Sites & Areas under active redevelopment with tower cranes, steel structures, and temporary facilities & Classified using ResNet model \\
        \cline{2-4}

        \multirow{3}{=}{\tikzmark{bottom}Redeveloped}
        & Buildings & Redeveloped buildings used for residential, commercial, or office purposes & Assigned based on POIs and OSM overlays \\
        \cline{2-4}
        & Green Spaces & Public open space, landscaped parks, or ecological green land & Assigned based on POIs and OSM overlays \\
        \cline{2-4}
        & Others & Other built-up land, such as roads, plazas, public service facilities, parking areas, or stadiums, etc. & Assigned based on POIs and OSM overlays \\
        \bottomrule
    \end{tabular}
    }

    \begin{tikzpicture}[remember picture, overlay]
     \draw[-{Latex[length=2.5mm]}, thick]
        ([xshift=14pt, yshift=-30pt]pic cs:top) -- 
        ([xshift=14pt, yshift=0pt]pic cs:mid);

     \draw[-{Latex[length=2.5mm]}, thick]
        ([xshift=14pt, yshift=-25pt]pic cs:mid) -- 
        ([xshift=14pt, yshift=25pt]pic cs:bottom);
    \end{tikzpicture}

\end{table}

Specifically, \textit{incomplete demolition} refers to parcels significantly reduced in size by $T_3$, yet not completely cleared. Vacant land and construction sites, typically lacking explicit POIs or OSM tags, are identified using spectral and textural features derived from remote sensing imagery. \textit{Vacant land} generally displays homogeneous bare surfaces with sparse vegetation \citep{mao_large-scale_2022}, whereas \textit{construction sites} exhibit irregular reflectance patterns accompanied by machinery and temporary structures \citep{wu_improvement_2024}. To accurately classify these transitional land-use types, we employ a ResNet model \citep{he_deep_2015}, which is known for its effective extraction of deep image features due to its residual learning architecture.

For redeveloped parcels, semantic assignment is conducted through the integration of POIs data and OSM layers. Residential, commercial, or office uses are classified as \textit{buildings}; parks and landscaped areas as \textit{green spaces}; and roads, plazas, or public infrastructure as \textit{others}. Notably, parcels demolished from UVs rarely convert directly into single-use categories but often evolve into mixed-use spaces. To ensure accuracy, we implement detailed spatial clipping for these parcels. In areas with overlapping categories, the classification relies on the dominant pixel-based land-use type, supplemented by manual verification to maintain spatial exclusivity and semantic consistency.

Following classification, we quantify the area associated with each land use type and develop a transition matrix to examine spatiotemporal dynamics. These metrics reveal how clearance and redevelopment processes differ across cities and offer insights into broader patterns of land restructuring in post-demolition contexts. 

\section{Results and analysis}
\label{sec4}
\subsection{Urban village mapping}
\label{sec4.1}
\subsubsection{Model performance evaluation}
\label{sec4.1.1}
Table \ref{tab:metrics_table} summarizes the segmentation performance of three models across the four cities. UV-SAM outperformed its counterparts in Zhengzhou and Harbin, achieving the highest mIoU scores of 90.30 and 85.77, respectively. It also recorded superior UV-IoU and UV-Precision, reflecting robust delineation of UV boundaries and effective suppression of false positives. 

\begin{table}[h!]
    \centering
    \caption{UV identification performances across datasets and methods. The best results are highlighted in bold.}
    \label{tab:metrics_table}
    \resizebox{\textwidth}{!}{
    \begin{tabular}{c c|c c c c|c}
        \hline
        \textbf{Dataset} & \textbf{Method} & \textbf{mIoU} & \textbf{UV-IoU} & \textbf{UV-Recall} & \textbf{UV-Precision}  &\textbf{Mixed Cities-mIoU}\\ \hline
        
        \multirow{3}{*}{Guangzhou}& U-Net          & 84.76 & 76.55  & 82.71 & \textbf{91.14}  & 83.14\\ 
            & DeepLab-v3+     & \textbf{85.28} & \textbf{77.54} & \textbf{86.01} & 88.73  & 84.03\\ 
            & UV-SAM          & 71.11& 71.17& 70.20& 82.95 & 71.01\\ \hline
        \multirow{3}{*}{Zhengzhou}& U-Net          & 88.16& 82.99& 91.09& 90.32& 86.26\\ 
           & DeepLab-v3+     & 88.37& 83.33& \textbf{91.98}& 89.87& 86.51\\ 
            & UV-SAM& \textbf{90.30} & \textbf{83.96} & 90.23 & \textbf{90.92}  & 88.34\\ \hline
            \multirow{3}{*}{Xi'an} 
            & U-Net          & 83.03 & 72.00& 78.32 & \textbf{89.93}  & 81.52\\ 
            & DeepLab-v3+     & 82.86 & 71.92  & \textbf{80.71} & 86.84  & 82.34\\ 
            & UV-SAM         & \textbf{84.07} & \textbf{73.80} & 79.29 & 87.02  & 83.97\\ \hline
 \multirow{3}{*}{Harbin}& U-Net          & 84.81& 77.97& 83.65&\textbf{92.00} & 83.30\\
 & DeepLabv3+     & 82.55& 76.25& \textbf{86.57}&90.24 & 81.59\\
 &  UV-SAM         & \textbf{85.77}& \textbf{79.31}& 83.91&86.36 & 83.28\\ 
  \hline
    \end{tabular}
    }
\end{table}

In Xi’an, UV-SAM yielded high precision but slightly lower recall than DeepLab-v3+, suggesting reduced sensitivity to fragmented or peripheral UV patterns. In contrast, DeepLab-v3+ performed best in Guangzhou, achieving an mIoU of 85.28 and the highest UV-Recall (86.01), demonstrating better coverage in densely built environments. U-Net, while showing the highest UV-Precision in Harbin (92.00) and Guangzhou (91.14), consistently lagged in UV-IoU and recall, indicating its limited ability to capture complete UV extents. 

City-level comparisons further reveal that Zhengzhou exhibited the most balanced performance across all models, reflecting clear UV boundaries and high-quality training data. Harbin also showed strong results, particularly under UV-SAM. Conversely, Xi’an recorded the lowest overall accuracy, likely due to image quality, complex urban morphology, or shadow interference. In Guangzhou, while recall was high, the model tended to over-predict in dense areas, leading to slightly lower precision.

Notably, models trained on Mixed-Cities datasets consistently underperformed their city-specific counterparts, with mIoU scores typically 0.1 to 2.5 points lower. This underscores current limitations in cross-city generalization and reinforces the necessity of localized training strategies when mapping UVs with pronounced morphological diversity \citep{yu2025exploring}. 

\subsubsection{Spatiotemporal dynamics of urban village extent }
\label{sec4.1.2}
\begin{figure}
    \centering
    \includegraphics[width=1\linewidth]{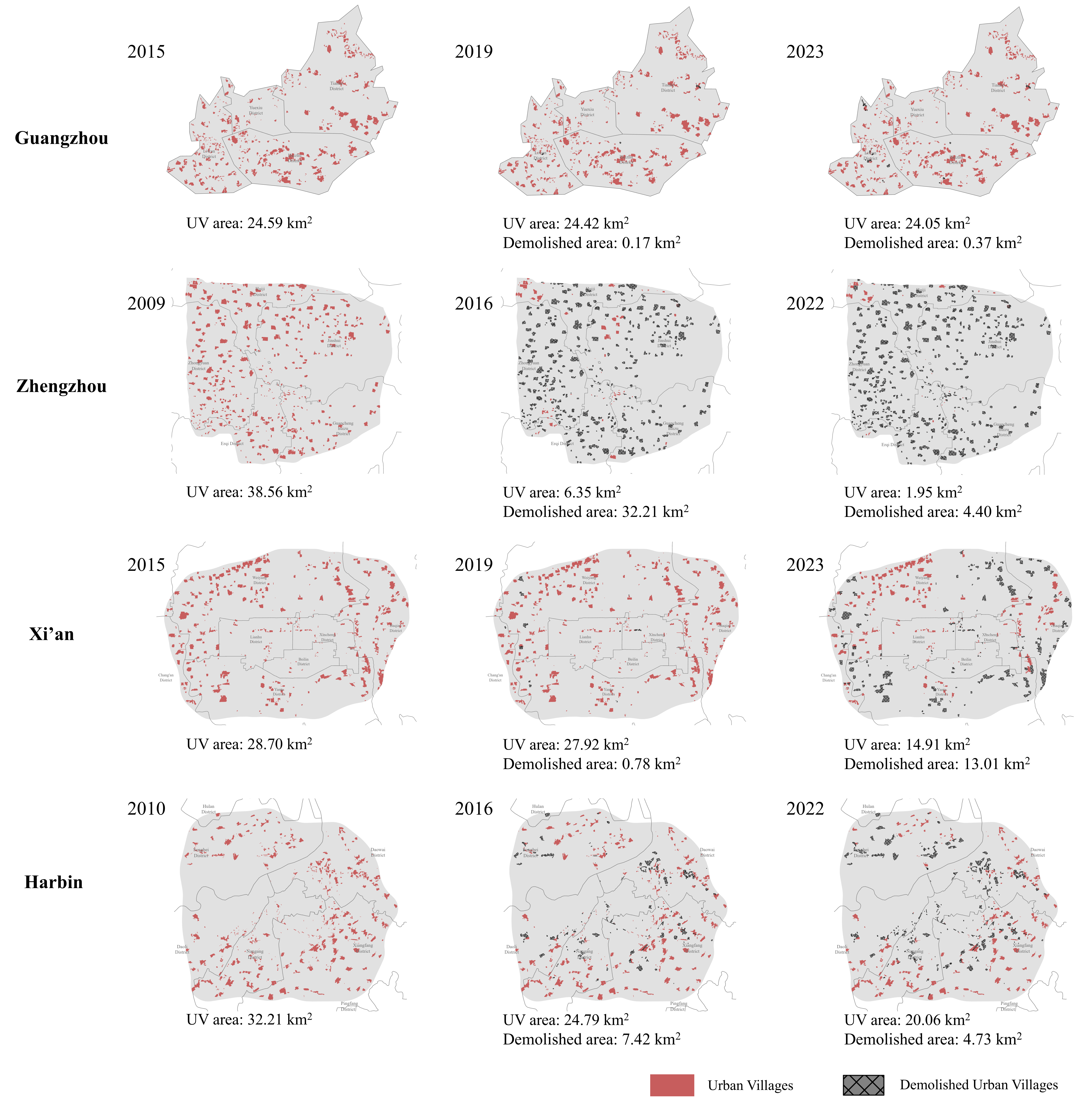}
    \caption{Spatiotemporal distribution of urban villages in the four cities. The results are derived from post-processed multi-temporal segmentation. Urban villages are denoted in red, \revb{while demolished sites are marked with hatching, overlaid on urban administrative boundaries (gray)}. \revb{The total UV area and the corresponding demolished area relative to the prior year are indicated below each map.}}
    \label{fig:UV_results}
\end{figure}

Figure \ref{fig:UV_results} illustrates the spatiotemporal dynamics of UV distribution in the study area. While all four cities experienced a reduction in total UV area, the pace and magnitude of decline varied substantially. 
In Guangzhou, the UV area decreased only slightly from 24.59~km² in 2015 to 24.05~km² in 2023. This modest change suggests a gradual and adaptive redevelopment strategy that likely aims to minimize disruptions to local communities and economic activity. In contrast, Zhengzhou underwent the most substantial transformation. The UV area declined from 38.56~km² in 2009 to 6.35~km² in 2016, and further to 1.95 km² by 2022. This reduction of over 90 percent within thirteen years indicates a strong emphasis on concentrated clearance and extensive spatial restructuring. 

Xi’an and Harbin exhibited distinct temporal patterns. In Xi’an, UV transformation progressed slowly in the early stage, with area decreasing from 28.70~km² in 2015 to 27.92~km² in 2019. By 2023, however, the area had declined sharply to 14.91~km², representing nearly a 50\% reduction compared to 2015. Harbin, by contrast, experienced more rapid early reductions, with UV area shrinking from 32.21~km² in 2010 to 24.79~km² in 2016, followed by a slower decrease to 20.06~km² by 2022. 

These results highlight not only the varying pace and extent of UV redevelopment across cities, but also prompt a critical question: what types of land use have replaced the demolished UVs, and to what extent have these areas been reintegrated into the broader urban fabric? The following section addresses this issue in detail. 

\subsection{Land-use transformation of urban villages }
\label{sec4.2}
\subsubsection{Overview of land-use changes after urban village demolition}
\label{sec4.2.1}
Demolished and incomplete demolition parcels of UVs were identified through spatial differencing of multi-temporal UV boundaries. Redeveloped categories, including buildings, green spaces, and others, were determined through the overlay of POIs and OSM data. Vacant land and construction sites, as transitional types, were classified using a ResNet18 model trained on 250 manually labeled image patches, achieving F1-scores of 0.94 and 0.88, respectively.

\begin{figure}
    \centering
    \includegraphics[width=1\linewidth]{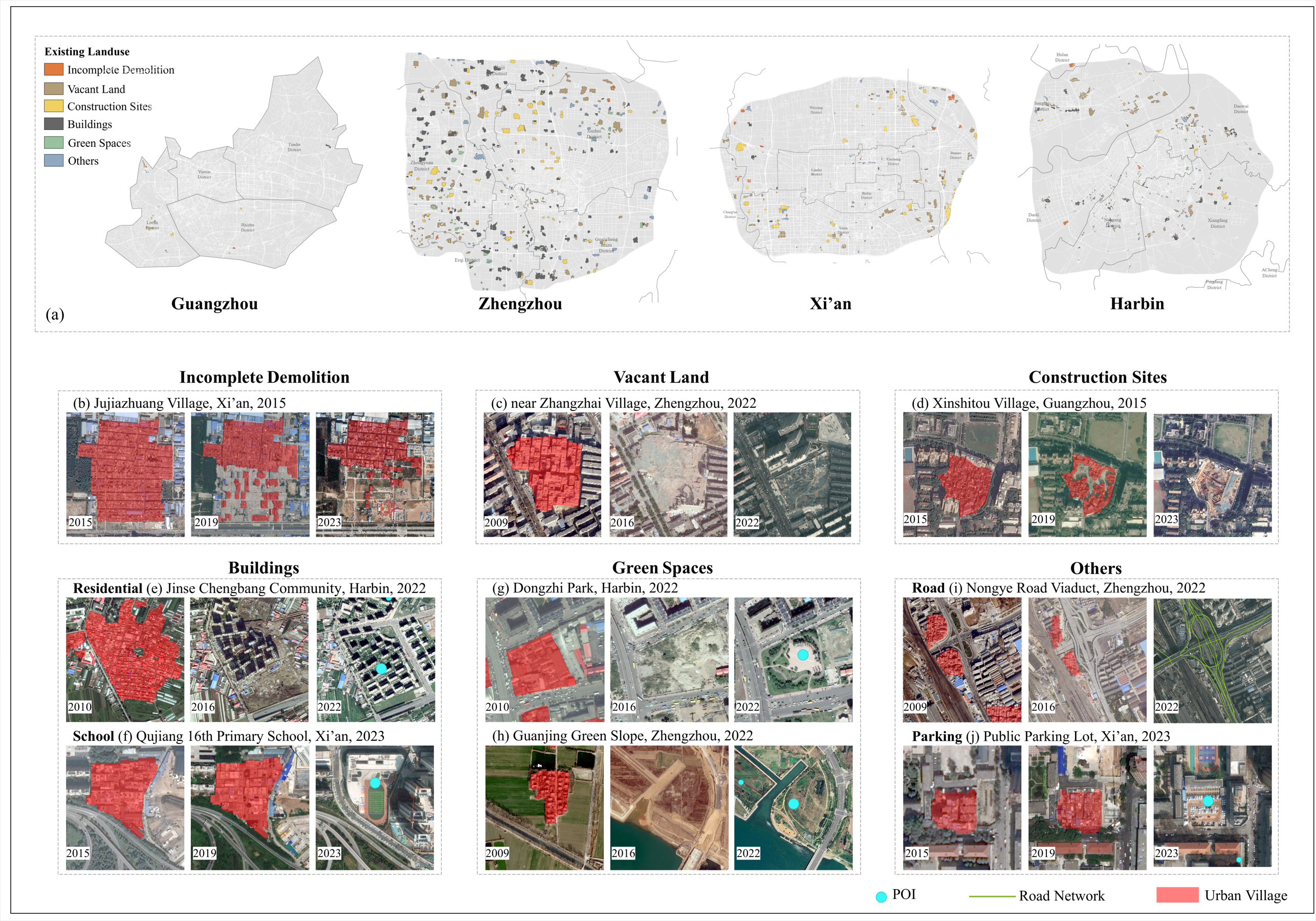}
    \caption{Spatial distribution and typical examples of different land use types after demolition of urban villages. (a) Spatial distribution of land use transitions from demolished UVs, categorized by type and overlaid with road networks. (b-j) Representative cases of different transformation types across years.}
    \label{fig:ChangeType}
\end{figure}

At the city scale, as shown in Figure \ref{fig:ChangeType}(a), redevelopment trajectories of UVs demonstrate strong spatial heterogeneity. In cities with a longer observation window (2009–2022), Zhengzhou features high-intensity redevelopment, though many parcels remain underutilized; Harbin also implemented substantial redevelopment, but its northeastern outskirts still contain large vacant zones, indicating delayed spatial stabilization. In contrast, cities with shorter timeframes (2015–2023) present distinct strategies: Xi’an concentrated redevelopment along the second and third ring roads, with a high proportion of construction land, reflecting synchronized demolition and rebuilding. In contrast, Guangzhou pursued small-scale, dispersed redevelopment within the urban core, reflecting an incremental and adaptive renewal strategy. Notably, across all four cities, many of the demolished UV sites coincide spatially with newly constructed road networks, indicating that clearance activities have contributed to the expansion of transport infrastructure and the reconfiguration of urban spatial structures, thereby improving inter-neighborhood connectivity. 

Figure \ref{fig:ChangeType}(b–j) present representative cases of current land uses on previously demolished UV sites across the study cities, tracing their respective transformation pathways. Most cases exhibit a sequential progression—from incomplete demolition, through vacancy or construction phases, to eventual redevelopment. However, the pace, allocation, and spatial configuration of these transformations vary considerably among cities, shaped by differences in governance capacity, planning orientation, and market dynamics.
In the early stage, some sites remain in a state of incomplete demolition. For instance, the Jujiazhuang site in Xi’an (Figure 5.b) showed signs of incomplete demolition as early as 2019, yet remained substantial building remnants by 2023, indicating stalled redevelopment. Fully demolished sites generally fall into two transitional categories: vacant and construction sites. Zhangzhai Village in Zhengzhou (Figure 5.c) remained vacant between 2016 and 2022, suggesting delayed follow-up investment; in contrast, Xishitou Village in Guangzhou (Figure 5.d) entered the construction phase promptly, reflecting a clear development path and high implementation efficiency.

Several sites have been fully transformed into developed urban functions. In Harbin and Xi’an, the Jinse Chengbang Community and Qujiang No. 16 Primary School (Figure 5.e, 5.f) exemplify conversion into residential and educational spaces. Harbin’s Dongzhi Park and Zhengzhou’s Guangjing Green Slope (Figure 5.g, 5.h) highlight the ecological direction of land redevelopment. Infrastructural redevelopment is illustrated by the Nongye Elevated Road in Zhengzhou and a public parking facility in Xi’an (Figure 5.i, 5.j).

\subsubsection{Quantitative analysis of urban village transformation}
\label{sec4.2.2}
\begin{landscape}
\begin{figure}
    \centering
    \includegraphics[width=1.4\textheight]{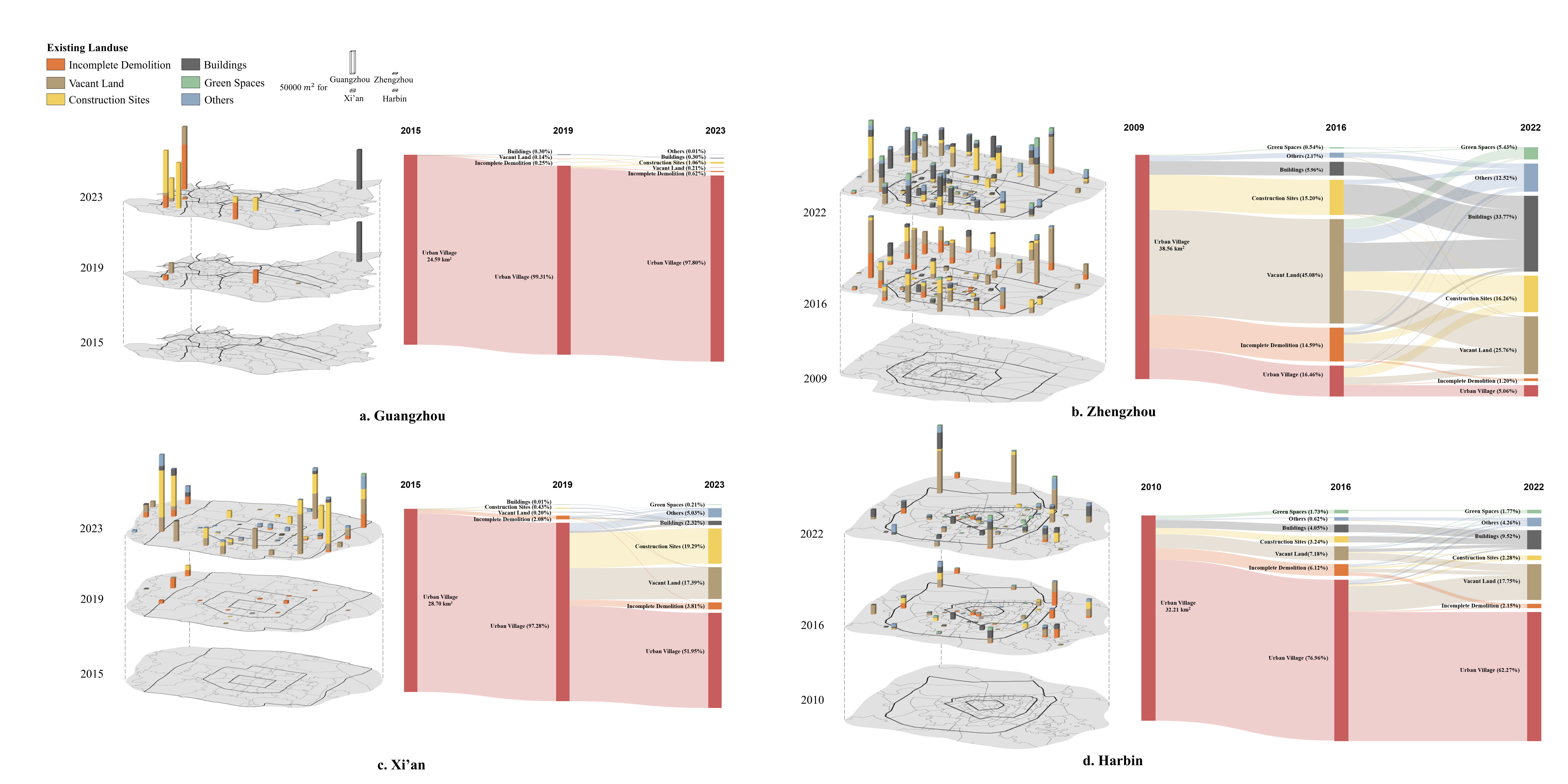} 
 \caption{Maps and Sankey diagrams of land use changes following the demolition of urban villages. The 3D bar charts visualize land-use transformations at the sub-district level, with bar height representing the area of land use change. The location of each bar corresponds to the centroid of the respective sub-district. Note that each city adopts a different vertical scale, as indicated in the legends. Thick black lines in the base maps denote major urban ring roads. The Sankey diagrams on the right illustrate the temporal flow of land use transitions, where the initial value corresponds to the total urban village area in the baseline year ($T_1$).}
    \label{fig:sanky}
\end{figure}
\end{landscape}

To analyze post-demolition transformations of UVs, we generated land use transition maps at the sub-district (Jiedao) level across the four cities (Figure \ref{fig:sanky}). Combining Sankey diagrams with 3D bar visualizations, the results reveal notable divergences in redevelopment tempo, spatial response, and functional transition pathways. 

\textbf{Redevelopment Pace.} Zhengzhou and Harbin underwent early large-scale demolition, reducing the share of UV area to 5.06\% and 62.27\%, respectively. However, persistently high vacancy rates (25.76\%, 17.75\%) suggest a significant gap between clearance and redevelopment. Xi’an exhibited stronger coupling between demolition and construction, with the proportion of UVs dropping to 51.95\% and over 20\% of land transitioning to building or construction use. In contrast, Guangzhou maintained relatively stable share of UV area and vacancy rates below 1\%, reflecting a restrained, low-intensity approach. 

\textbf{Spatial Patterns.} All cities followed a common pattern of core stability and peripheral transformation. Most changes occurred beyond the third ring road. Zhengzhou and Harbin developed extensive vacant and construction zones along the urban fringe—indicative of inefficient development belts. Xi’an showed coordinated redevelopment along transportation corridors, while Guangzhou’s changes were more scattered and localized, with minimal disruption to the existing urban fabric. 

\textbf{Functional Transitions.} Cities also diverged in land-use priorities. Xi’an prioritized construction, with limited ecological or service-oriented reallocation. Zhengzhou remained 33.77\% of the land for building, while allocating 5.43\% to green spaces and 12.52\% to other infrastructure; 16.26\% remained under construction. Harbin increased ecological space but showed limited construction momentum. Guangzhou preserved its original land-use pattern, indicating the mildest transformation. 

The analysis results and visualizations underscore that UV redevelopment is a prolonged and often fragmented process, with few cleared areas achieving stable land use even after four to six years—highlighting the delayed spatial stabilization following demolition. This pattern reflects the nonlinear and asynchronous nature of UV transformation, where differences in redevelopment pace, spatial configuration, and functional restructuring reveal varying capacities for resource absorption, market activation, and policy execution. Particularly under large-scale demolition scenarios, spatial clearance frequently outpaces land reuse, posing challenges for long-term planning efficiency and urban equity.

\section{Discussion}
\label{sec5}
This study was motivated by the lack of systematic and scalable tools to monitor the vanishing and transformation of UVs amid China’s shift toward more sustainable and inclusive urban development. By leveraging multi-temporal satellite imagery and deep learning techniques, we developed a framework to identify, quantify, and compare UV redevelopment across four mega-cities. This section discusses the methodological reliability, spatial and functional transition patterns, and broader policy implications, and proposes directions for more socially responsive urban renewal.

\subsection{Accuracy and reliability of the dynamic urban village mapping results}
\label{sec5.1}
The generalizability and adaptability of deep learning models remain critical challenges in the detection of informal settlements, especially within complex urban environments. This study conducted empirical evaluations across four major Chinese mega-cities representing different economic regions and urban contexts. Among the three segmentation models tested, UV-SAM exhibited superior performance in most cases, although its accuracy was relatively lower in Guangzhou. This suggests that current algorithms still face limitations in adapting to heterogeneous UV morphologies and spatial textures \citep{yu2025exploring}.

\begin{figure}
    \centering
    \includegraphics[width=1\linewidth]{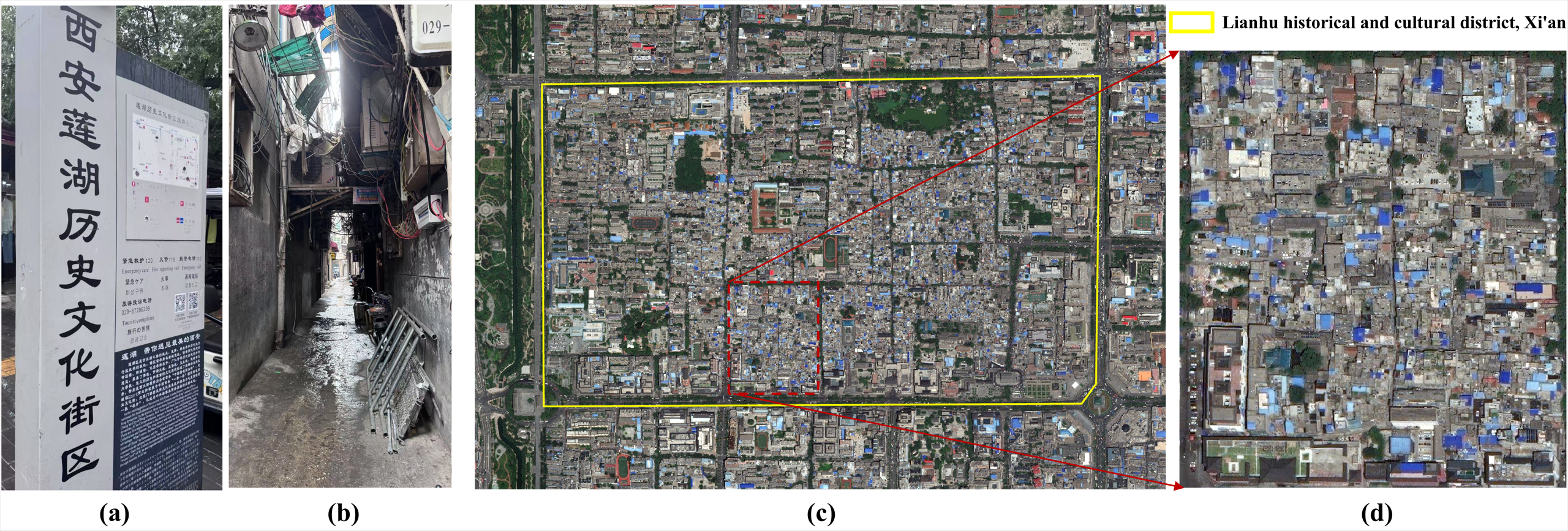}
    \caption{Example of a government-designated historical and cultural conservation district in Xi’an. (a) entrance signage indicating the protected status. The Chinese characters means Lianhu Historical and Cultural District, Xi’an; (b) a typical alleyway reflecting dense and low-rise urban morphology; (c) satellite imagery highlighting the district boundaries (in yellow); and (d) a close-up view revealing compact housing structures with irregular layouts. Such districts often resemble urban villages in morphology, posing challenges for automatic classification models. 
    (Data source: Street view photographs by the author, June 2024; satellite imagery from Google Earth, 2023.)}
    \label{fig:historical districts}
\end{figure}

Another significant challenge lies in distinguishing UVs from historical and cultural districts, which often share similar spatial characteristics such as high density, low-rise structures, and irregular morphology. Figure \ref{fig:historical districts} shows a typical example from the Lianhu Historical and Cultural District in Xi’an. Previous studies have noted that models sometimes misclassify these areas as UVs, and attributed such errors to the limited quality of satellite imagery \citep{zhangUVSAMAdaptingSegment2024}. To avoid such misidentification, we excluded all officially designated heritage zones based on local government planning boundaries \footnote{Official sources for government-designated heritage boundaries in Guangzhou (\url{https://zfcxjst.gd.gov.cn/zwzt/pzts/bhcc/gzjl/content/post_4182807.html}), Zhengzhou (\url{https://public.zhengzhou.gov.cn/D1102X/234264.jhtml}), Xi’an (\url{https://www.shaanxi.gov.cn/zfxxgk/fdzdgknr/zcwj/nszfwj/szh/202208/t20220808_2241979_wap.html}), and Harbin (\url{http://hlj.people.com.cn/n2/2023/1109/c220005-40634116.html}).} and extensive field verification. This refinement significantly enhanced the semantic consistency and policy relevance of our results. The findings highlight that improving classification accuracy requires not only algorithmic refinement, but also a nuanced understanding of the planning semantics embedded in local spatial governance. 
Furthermore, although our model did not detect large-scale UV demolition in Guangzhou, this should not be interpreted as a lack of urban renewal activities. As a national pioneer in incremental regeneration, Guangzhou has prioritized micro-scale improvements and community-level upgrading over wholesale clearance \citep{ZRZX202501016}. These subtle transformations are often difficult to capture through conventional satellite imagery, indicating the need for future research to integrate additional features, such as building height, roof material, and street-level visual data, to better detect and interpret fine-grained urban changes. 

To analyze post-demolition redevelopment, we developed a land-use classification schema grounded in the degree of utilization rather than conventional functional zoning \citep{caoDeepLearningbasedRemote2020}. Instead of assigning plots to fixed categories such as residential or commercial, we delineated three progressive stages—remained, demolished, and redeveloped—further divided into six subtypes: incomplete demolition, vacant land, construction site, building, green spaces, and others. This typology enables a more dynamic, process-oriented understanding of UV transformation and reflects both physical and functional changes over time.

Despite its advantages, this classification approach is subject to several uncertainties. First, semantic interpretation of land-use types is affected by the cultural and institutional contexts of different cities—what appears to be a vacant lot in Xi’an may, in fact, be under archaeological protection rather than representing inefficient land use. Second, the classification model relies heavily on publicly available datasets and visual interpretation, which may not always reflect real-time land-use transitions. Ambiguities frequently arise in transitional zones, such as those between green spaces and construction, or temporary structures and permanent buildings, increasing the interpretive burden and classification instability.

\subsection{From vanishing to transformation: a critical phase}
\label{sec5.2}
\subsubsection{Vanishing: displacement, housing loss, and urban inequality}
\label{sec5.2.1}
The vanishing of UVs is not merely a matter of physical erasure—it is the quiet unraveling of deeply rooted communities that once sheltered the city’s most mobile, yet most vulnerable, residents. For decades, UVs have provided essential housing for low-income and migrant populations in Chinese cities. However, despite their crucial social function, they have often been stigmatized due to the complex social issues they are seen to represent \citep{jin_destigmatizing_2024}. 

Under the framework of SDG 11, many cities have launched UV redevelopment initiatives. Yet ``vanishing'' goes beyond demolition, it marks a rupture in housing provision and institutional support. Without adequate resettlement and policy safeguards, such clearance risks amplifying spatial injustice and social exclusion. In cities like Zhengzhou, redevelopment has proceeded swiftly and at scale. As detailed in Section \ref{sec4.2}, mass demolitions were completed rapidly, yet functional replacements lagged significantly, leaving one-quarter of cleared sites vacant. These ``urban scars'' reflect both inefficient land use and the emergence of new social and safety risks \citep{garvin_more_2013}. 

More critically, the loss of UVs undermines affordable housing stock and disrupts migrant residents' social networks \citep{lin_go_2025,li_dawn_2021}. Displaced populations are often pushed to more peripheral or equally unstable settlements, perpetuating cycles of forced relocation and insecure tenure \citep{CSGH202002008,tian2020systems}. For example, according to \cite{yangguangwang70w}, Zhengzhou's demolition of UVs affected approximately 700,000 migrant workers. However, the combination of soaring housing prices, a shortage of public rental housing, and an increasingly tight rental market forced many migrants to relocate frequently \footnote{Fang.com’s investigative report on Zhengzhou urban village redevelopment (2016). \url{https://fdc.fang.com/news/2016-02-18/19684920.htm}}. Some rural migrants were ultimately pushed out of the city altogether. This undermines both social equity and labor accessibility. \revb{To further capture how these processes were perceived at the public level, we employed the official Weibo AI Search service \footnote{Weibo AI Search (powered by large language model \textit{DeepSeek R1}) \url{https://s.weibo.com/aisearch}} to analyze discussions on UV demolition in Zhengzhou (details can be found in Supplementary Appendix A.1). As one of China’s largest social media platforms with broad penetration across user groups, Weibo provides a reliable source for gauging public sentiment \citep{shi2022online}, and the AI Search service systematically retrieves and aggregates relevant posts. The results indicate that public discourse primarily centered on three themes: (1) the disappearance of low-rent housing following the intensive wave of demolitions around 2016; (2) long delays in the delivery of resettlement housing, which increased household financial burdens; and (3) the recent implementation of “housing voucher resettlement” schemes and the allocation of national redevelopment funds, which have sparked debates over insufficient compensation, limited housing choices, and extended resettlement cycles. These voices reflect the profound impacts of UV demolition on residents' lives. They also highlight blind spots in post-demolition processes, particularly the overlooked spaces that shape resettlement experiences.}

One critical overlooked aspect concerns the dynamics of informal resettlement sites that emerge following UV clearance. These areas, often excluded from formal planning frameworks, accommodate displaced migrant tenants and other non-owner residents who previously relied on UVs for affordable housing. As they are typically formed outside official relocation schemes, such spaces play a pivotal role in shaping the lived experiences and spatial trajectories of vulnerable populations. Mapping and monitoring these “hidden geographies” could offer critical insights for promoting equitable urban development and addressing the blind spots of post-demolition resettlement \citep{gao2018exploring}. 

\subsubsection{Transformation: delayed redevelopment and green resilience potential}
\label{sec5.2.2}
The transformation of UVs reflects broader shifts in urban dynamics rather than mere technical or logistical challenges. Historically, UVs emerged as a byproduct of China’s rapid urban expansion \citep{cao2025mapping}, bridging gaps between formal planning and actual urban growth. However, as many Chinese cities now confront slowing growth or even population decline \citep{zou_four_2025}, the underlying logic driving redevelopment is shifting. 

This transition is particularly evident in cities such as Harbin and Zhengzhou. As discussed in Section \ref{sec4.2.2}, although large-scale demolitions have been completed, many cleared sites remain either vacant or delayed in prolonged construction processes. In Harbin, vacant land constituted 47\% of all cleared UV parcels by 2022—the highest proportion among the four cities. This situation reflects broader structural constraints in Northeast China, where economic decline and demographic shrinkage have undermined both fiscal capacity and market confidence \citep{tong_understanding_2021}. In Zhengzhou, aggressive clearance campaigns were implemented early, yet limited follow-up planning and weak institutional coordination have resulted in widespread underutilization. These cases challenge the conventional assumption that demolition will naturally lead to timely redevelopment. UVs are increasingly suspended between the retreat of expansion-driven planning and the unrealized vision of renewal.

Despite these challenges, the transformation of UVs holds significant spatial potential. While some parcels are repurposed for housing, transportation, or utilities, integrating green spaces provides a cost-effective and flexible strategy for climate resilience \citep{ZHANG2025106576}. Green infrastructure has been shown to mitigate urban heat island effects, improve microclimates, and support public health and adaptive capacity \citep{wu_green_2024,RUI2024105122}. In this study, we classified green spaces as a separate category to quantify its proportion among transformed sites. Our findings suggest that several cities are proactively expanding ecological networks, signaling a more resilient approach to spatial redevelopment.

Importantly, many UV areas coincide with zones of heightened climate vulnerability \citep{elmarakby_prioritising_2024,chen_assessing_2025}. A comparison between identified UVs in Xi’an and the mapped distribution of urban heat island intensity and climate inequality \citep{xu_influences_2024} reveals substantial spatial overlap. In addition, inadequate drainage and poor waste management have made these areas highly susceptible to flooding \citep{sakijege_flooding_2012,chen_assessing_2025,balderas_torres_systemic_2020}. In cases of in-situ upgrading or incremental renewal, as opposed to wholesale demolition, strengthening basic infrastructure remains essential. Targeted improvements in drainage, green cover, and public space can significantly strengthen these communities’ capacity to cope with extreme weather events.

\subsection{Governance evolution and implementation gaps }
\label{sec5.3}
\begin{figure}
    \centering
    \includegraphics[width=1 \linewidth]{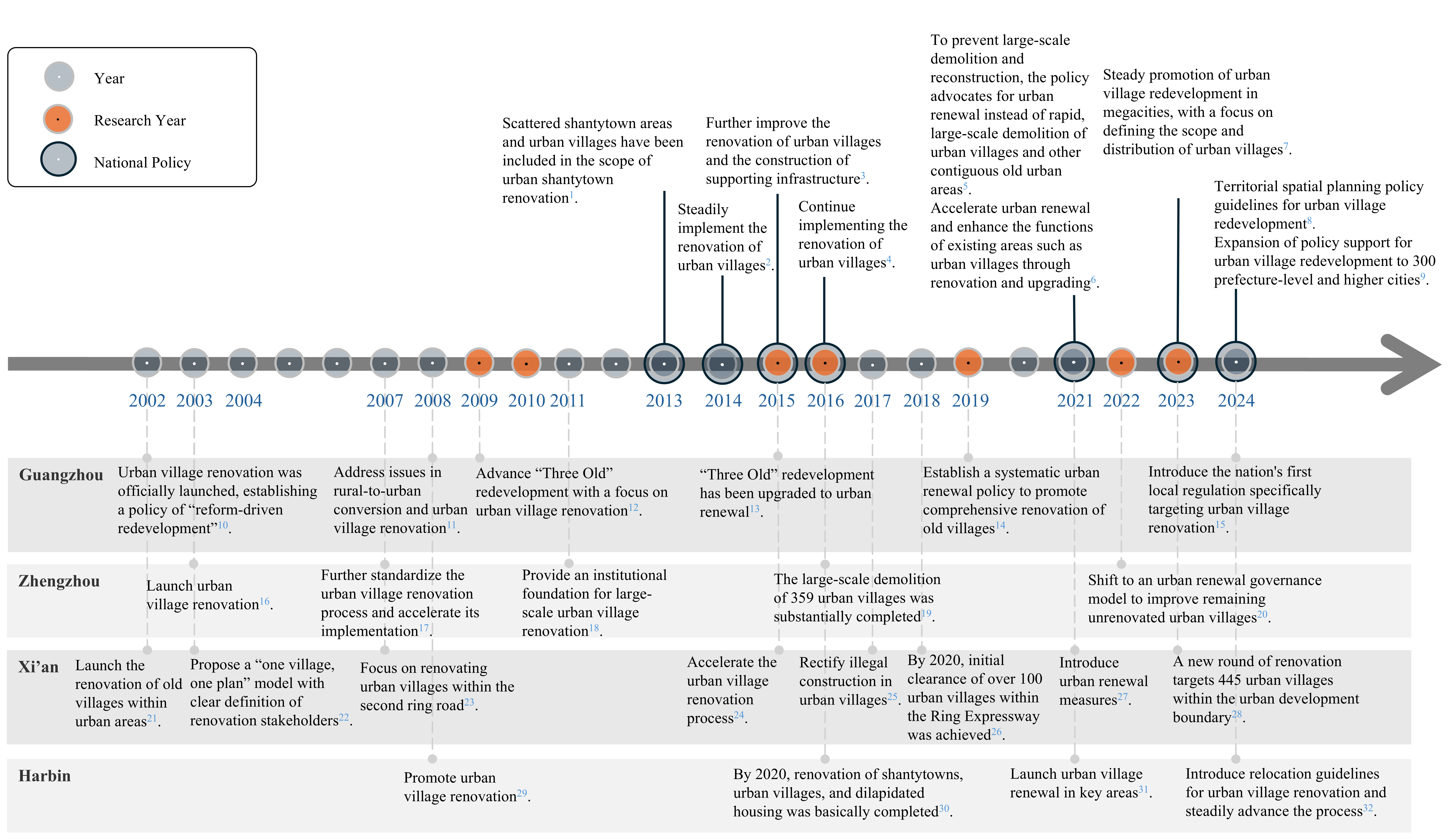}
    \caption{Timeline of national and local urban village redevelopment policies in China (2002–2024). 
Full policy document references are listed in Supplementary Appendix A.2 by sequence number.  } \label{fig:Policy}
\end{figure}

As highlighted earlier, UV redevelopment in China is deeply shaped by national and local policies that coordinate spatial planning, population resettlement, and infrastructure provision. Policy orientations not only guide implementation strategies but also shape the governance logics and spatial outcomes across different cities. This section traces the evolution of key policies from 2002 to 2024 (as shown in Figure \ref{fig:Policy}), highlighting a trajectory from land-driven restructuring toward more holistic approaches emphasizing sustainable urban renewal and territorial governance over the two decade's rapid urbanization in China. 

At the national level, the inclusion of UVs in the official shantytown redevelopment agenda in 2013 marked a turning point, elevating their governance to a priority issue. Since 2021, the policy direction shifted from large-scale demolition and reconstruction to urban renewal. This new approach emphasizes preservation, improvement, and organic transformation. In 2023 and beyond, the central government further clarified the spatial scope and distribution of UVs. It also expanded the renovation target to more than 300 prefecture-level cities, marking a new phase of area-wide UV governance.

At the city level, \revb{the policy shifts of UV redevelopment varied considerably across cities, and their} implementation trajectories vary considerably. \revb{Guangzhou was among the earliest cities to prioritize the urban renewal of UVs, upgrading the} “Three-Old” redevelopment strategy (targeting old towns, villages, and factories) \revb{to urban renewal in 2015. Between 2015 and 2023, the spatial patterns of Guangzhou’s UVs remained relatively stable.} In 2024, the city enacted the Regulations on UVs Renovation, institutionalizing its micro-renewal strategy through a formal legal framework. In contrast, Zhengzhou pursued a high-intensity clearance model. \revb{In 2016, Zhengzhou announced plans for the large-scale demolition of approximately 350 UVs, but by 2022 had shifted toward an urban renewal governance model emphasizing the improvement of remaining villages. Between 2009 and 2016, the urban village area in Zhengzhou declined sharply, followed by a continued decrease from 2016 to 2022.} However, as shown in Sections \ref{sec4} and \ref{sec5.2.2}, this approach resulted in substantial tracts of idle land and delayed functional replacement, undermining resettlement effectiveness. \revb{In 2018, Xi’an set the goal of eliminating all UVs within its expressway ring by 2020, and in 2021 introduced urban renewal regulations. Between 2019 and 2023, the total area of UVs in Xi’an declined substantially. Owing to its distinctive historical legacy as an old industrial base, Harbin initiated shantytown redevelopment relatively early. In 2016, it announced plans to basically complete the redevelopment of shantytowns, UVs, and dilapidated housing by 2020, and in 2021 launched targeted renewal projects for key areas. Between 2010 and 2016, Harbin’s UVs decreased significantly, with the pace of reduction slowing between 2016 and 2022.}

Despite continued progress across cities, several implementation challenges remain. First, ambitious demolition targets often exceed the actual governance and fiscal capacity of local authorities, resulting in stalled or partial completion. Second, an overemphasis on land consolidation, fiscal returns, and development indicators tends to marginalize the needs of low-income residents, migrant workers, and other vulnerable groups. Current strategies often lack differentiated responses in resettlement, housing affordability, and community continuity, resulting in growing socio-spatial fragmentation and eroding urban inclusivity.

\subsection{Research limitations and future directions}
\label{sec5.4}
This study has several limitations that warrant further improvement. Technically, three key challenges remain. First, although the deep learning framework performs well in UV boundary mapping, it lacks end-to-end automation, which increases the difficulty in scaling the method to broader space and time. Second, while remote sensing captures large-scale morphological changes, it fails to reflect fine-grained functional attributes and governance details. Integrating additional data sources such as street view imagery could enhance spatial resolution and contextual accuracy. Third, no segmentation model proved universally effective due to significant morphological variation between northern and southern cities. Scaling to national-level monitoring is hindered by high computational demands and extended inference times. These issues may introduce errors in UV boundary mapping and post-demolition land use classification, yet still offer meaningful insights into broader spatial transformation trends. Additionally, the limited availability of VHR-RS imagery prevented full temporal alignment across the four study areas, which constrained the consistency of cross-city comparisons. Nonetheless, this constraint also enabled the study to incorporate both long-term (13 years) and short-term (8 years) observation windows, offering complementary temporal perspectives.

\revb{Despite these limitations, the proposed framework also has potential applications in other national contexts with informal settlements such as slums or favelas. While Chinese UVs differ in morphology and institutional context, informal settlements worldwide often share key physical features in remote sensing imagery, including high density, small roof sizes, irregular layouts, and sharp contrasts with surrounding formal areas \citep{kamalipour2019mapping}. Building on these shared characteristics, our technical pipeline integrates multi-temporal VHR-RS imagery, OSM and POI data, semantic segmentation models, and post-classification land-use analysis, which together offer a relatively low-cost and adaptable approach. In addition, the lifecycle perspective of “remained, demolished, and redeveloped” provides a broadly applicable analytical lens to understand both the vanishing and transformation of informal settlements. Admittedly, cross-national applications still face challenges such as variations in data availability, differences in the semantic interpretation of land-use transitions, and divergent governance contexts. Addressing these challenges requires incorporating locally specific data sources, calibrating classification schemes to reflect cultural and functional nuances, and fostering collaborations with local researchers and practitioners \citep{kohli2016uncertainty,rafieian2023gaps}. Future research should therefore empirically test and refine the framework in diverse Global South contexts to evaluate its robustness, scalability, and policy relevance for informal settlement governance. }

From a governance perspective, future efforts should center on three key shifts: systematic policy evolution, \revb{a deeper understanding of transformation drivers, } and tiered intervention mechanisms in the Chinese context. Amidst slowing real estate growth \citep{GHSI202408002}, transitioning from indiscriminate demolition toward socially regenerative approaches is imperative. This paradigm shift necessitates balancing spatial optimization with socioeconomic equity while leveraging urban re-planning opportunities to address demographic shifts and climate pressures. \revb{A comprehensive understanding of transformation drivers, including policy, economic, social, and demographic factors \citep{GAO2023102875,rafieian2023gaps}, is essential for designing effective and context-sensitive governance strategies. Future work should incorporate quantitative approaches to assess the relative influence of these factors across different spatial and institutional contexts.} To support targeted action, we advocate establishing a tiered classification system based on locational attributes, functional typology, cultural significance, and resident demographics to categorize informal settlements into \textit{demolition-redevelopment}, \textit{micro-regeneration}, and \textit{preservation-enhancement}. This approach clarifies intervention priorities and sequencing, thereby improving policy precision and resource allocation.

These coordinated technical and governance advancements ultimately aim to achieve dual objectives: optimizing spatial structures while stabilizing social foundations, thus steering urban renewal toward high-quality, sustainable, and human-centered development. 

\section{Conclusion}
\label{sec6}
Amidst global efforts to improve informal settlements under the United Nations' SDG 11 and China’s renewed emphasis on UV renovation, this study offers a critical and cross-regional reflection on UV redevelopment practices in China. It provides valuable insights into the transformation patterns and associated land-use changes.

Specifically, by integrating high-resolution remote sensing imagery with deep learning–based segmentation models, we systematically mapped the disappearance and redevelopment of UVs nearly fifteen years across four representative cities in China’s major economic regions, i.e., Guangzhou (East), Zhengzhou (Central), Xi’an (West), and Harbin (Northeast). The analysis identifies three predominant land-use transition pathways following demolition: synchronized redevelopment, delayed redevelopment, and gradual optimization. Each pattern reflects differing spatial strategies and governance capacities. These findings underscore the mismatch between demolition cycles and actual land redevelopment, which can lead to spatial inefficiencies and increase the risk of social displacement. Furthermore, we have established a scalable and transferable framework for monitoring the dynamics of urban informal settlements through multi-source geospatial data. It also provides planning insights that support context-sensitive strategies to balance spatial restructuring with social equity, especially in the context of slowing urban expansion and rising demands for sustainable reconstruction.

Future research can incorporate multi-perspective datasets with higher spatial and temporal resolution to explore the socioeconomic and environmental impacts of UV redevelopment. A tiered governance approach can help enhance the effectiveness, resilience, and equity of urban renewal. Collectively, these contributions not only inform policy development in China but also offer globally relevant knowledge for managing the redevelopment of informal settlements.

\section*{Acknowledgments}
This work is supported by the Guangzhou-HKUST(GZ) Joint Funding Program (No. 2025A03J3639), the Guangdong Provincial Project (No. 2023QN10H717), and the HKUST(GZ) Undergraduate Research Program (No. UP2024U027).

\section*{Declaration of generative AI and AI-assisted technologies in the writing process}
During the preparation of this work the authors used ChatGPT for language polishing. After using this tool, the authors reviewed and edited the content as needed and take full responsibility for the content of the publication.

\bibliographystyle{elsarticle-harv}
\bibliography{references}

\end{document}